\documentclass{article}

\usepackage{microtype}
\usepackage{graphicx}
\usepackage{subfigure}
\usepackage{booktabs} 
\usepackage[ruled,vlined,algo2e]{algorithm2e}
\SetNlSty{}{}{\scriptsize} 
\usepackage{multirow}
\usepackage{booktabs}
\usepackage{hyperref}

\usepackage[accepted]{icml2025}

\usepackage{amsmath}
\usepackage{amssymb}
\usepackage{mathtools}
\usepackage{amsthm}

\usepackage[capitalize,noabbrev]{cleveref}

\theoremstyle{plain}

\theoremstyle{definition}

\theoremstyle{remark}

\usepackage[textsize=tiny]{todonotes}

\icmltitlerunning{Step Back to Leap Forward: Self-Backtracking for Boosting Reasoning of Language Models}

\begin{document}

\twocolumn[
\icmltitle{Step Back to Leap Forward: Self-Backtracking for Boosting\\ Reasoning of Language Models}



\icmlsetsymbol{equal}{*}

\begin{icmlauthorlist}
\icmlauthor{Xiao-Wen Yang}{nju,njuai}
\icmlauthor{Xuan-Yi Zhu}{nju,njuai}
\icmlauthor{Wen-Da Wei}{nju,njuai} 
\icmlauthor{Ding-Chu Zhang}{nju,njuai}
\icmlauthor{Jie-Jing Shao}{nju}\\
\icmlauthor{Zhi Zhou}{nju}
\icmlauthor{Lan-Zhe Guo}{nju,njusz}
\icmlauthor{Yu-Feng Li}{nju,njuai}
\begin{tabular}{c}
\textsuperscript{1} National Key Laboratory for Novel Software Technology, Nanjing University, China.\\
\textsuperscript{2} School of Artificial Intelligence, Nanjing University, China\\
\textsuperscript{3} School of Intelligence Science and Technology, Nanjing University, China\\
    \texttt{\{yangxw,zhuxy,weiwd,zhangdc,shaojj,zhouz,guolz,liyf\}@lamda.nju.edu}
   \end{tabular} 
\end{icmlauthorlist}

\icmlaffiliation{nju}{National Key Laboratory for Novel Software Technology, Nanjing University, China}
\icmlaffiliation{njuai}{School of Artificial Intelligence, Nanjing University, China}
\icmlaffiliation{njusz}{School of Intelligence Science and Technology, Nanjing University, China}


\icmlcorrespondingauthor{Lan-Zhe Guo}{guolz@lamda.nju.edu.cn}
\icmlcorrespondingauthor{Yu-Feng Li}{liyf@lamda.nju.edu.cn}

\vskip 0.3in
]

\begin{abstract}
The integration of slow-thinking mechanisms into large language models (LLMs) offers a promising way toward achieving  Level 2 AGI \textit{Reasoners}, as exemplified by systems like OpenAI’s o1. However, several significant challenges remain, including inefficient overthinking and an overreliance on auxiliary reward models. We point out that these limitations stem from LLMs’ inability to internalize the search process, a key component of effective reasoning. A critical step toward addressing this issue is enabling LLMs to autonomously determine when and where to backtrack, a fundamental operation in traditional search algorithms. To this end, we propose a self-backtracking mechanism that equips LLMs with the ability to backtrack during both training and inference. This mechanism not only enhances reasoning ability but also efficiency by transforming slow-thinking processes into fast-thinking through self-improvement. Empirical evaluations demonstrate that our proposal significantly enhances the reasoning capabilities of LLMs, achieving a performance gain of over 40\% compared to the optimal-path supervised fine-tuning method. We believe this study introduces a novel and promising pathway for developing more advanced and robust \textit{Reasoners}.
The code is available at \url{https://github.com/LAMDASZ-ML/Self-Backtracking}.
\end{abstract}

\section{Introduction}
\footnotetext{
OpenAI has proposed five steps towards AGI: \textit{Chatbots}, \textit{Reasoners}, \textit{Agents}, \textit{Innovators}, and \textit{Organizations}.
}
The incorporation of slow-thinking mechanisms has become a pivotal path for large language models (LLMs) to attain Level 2 AGI \textit{Reasoners} \footnotemark. 
Notable advancements in this domain include OpenAI’s o1 and o3 models \cite{o1}, which have spurred the development of deep-thinking models such as DeepSeek R1 \cite{guo2025deepseek} and Qwen QwQ \cite{qwq}. Representative techniques employ reinforcement learning to autonomously acquire deep thinking capabilities, enabling LLMs to engage in self-reflection and self-correction, analogous to the search mechanism.

\begin{figure}[t]
    \centering
    \includegraphics[width=0.50\textwidth]{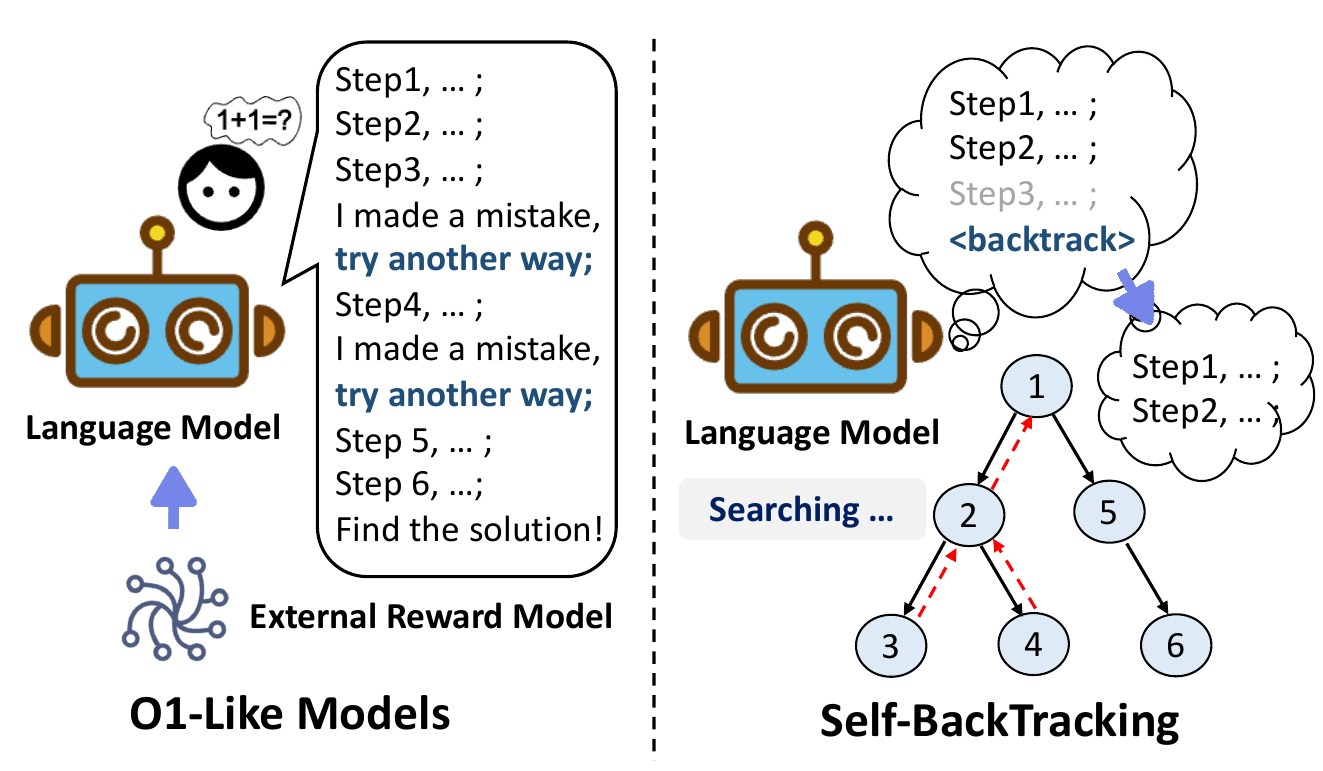}
    \caption{
       Comparison between o1-like models and our self-backtracking approach. O1-like models are prone to inefficient overthinking and exhibit overreliance on reward model.
        }
    \label{intro}
    \end{figure}
Nevertheless, do these techniques represent the ultimate paradigm for \textit{Reasoners}? Our analysis uncovers several critical challenges in current approaches: 1) O1-like models frequently suffer from inefficient overthinking, even for simple problems \cite{overthink}, resulting in considerable computational resource wastage; 2) Some o1-like methods \cite{choi2023kcts,zhang2024rest,qin2024o1} exhibit a heavy reliance on auxiliary reward models for state evaluation, which not only incurs significant inefficiencies but also risks reward hacking \cite{chen2024odin}.

The core issue lies in the fact that \textit{LLMs have yet to internalize the capability of search}. Currently, search functions more as an external component \cite{sos,zhang2024rest}, integrated with LLMs only at a superficial level, with limited fusion between the two. 
This externalized approach constrains the performance of LLMs in more complex reasoning tasks, while simultaneously inducing excessive thinking in simpler tasks. As the saying goes, “To err is human, to backtrack divine.” Backtracking \cite{van2006backtracking} plays a critical role in many efficient search algorithms, particularly in reasoning problems, where most classical search algorithms employ backtracking strategies. Through backtracking, algorithms can learn from suboptimal paths, reassess, and seek optimal solutions. Therefore, we have strong reasons to believe that if LLMs could internalize the backtracking mechanism, they would potentially mitigate overthinking and reduce reliance on external reward models, paving the way toward becoming stronger \textit{reasoners}.

In this paper, we propose a novel \textbf{Self-Backtracking} technique that equips language models with the ability to learn when and where to perform backtracking during the training phase. Specifically, the model is trained to recognize when its initial predictions or reasoning paths are suboptimal and to backtrack to earlier states in order to explore alternative possibilities. During the testing phase, the model leverages this learned backtracking capability to conduct dynamic search processes, systematically revisiting prior actions and exploring multiple reasoning paths. The model ultimately utilizes the improved results for expert iteration to achieve self-improvement. This process facilitates a transition from slow thinking to fast thinking, significantly enhancing the model's capability for fast reasoning in complex tasks. 

Our proposed method effectively addresses the limitations of o1-like models through the internalization of the backtrack process within the LLMs.
Firstly, the model intelligently avoids unnecessary backtracking in simpler problems by learning optimal conditions for backtracking, thereby effectively mitigating the risk of overthinking. Secondly, it implicitly integrates the state evaluation mechanism within the model itself, obviating the need for external reward models. 
 We compare o1-like models and our method in Figure \ref{intro}. Experiments are conducted on the Countdown task \cite{sos} to evaluate our proposed method’s advantages across models of different parameter scales. 
Our method demonstrates an accuracy enhancement exceeding 40\% compared to the SFT approach that solely relies on the optimal reasoning solutions.
Results show that the self-backtracking technique significantly enhances the model’s reasoning flexibility and overall performance while exhibiting test-time scaling capabilities. We believe it provides the potential to make a advancement toward achieving Level 2 AGI \textit{Reasoners}.

Our contributions are summarized as follows:
\begin{itemize}
    \item \textbf{Problem}: Existing slow thinking techniques face significant challenges, including inefficient overthinking and excessive reliance on auxiliary reward models. We highlight that enabling LLMs to internalize the search process is a promising direction for enhancing LLM's reasoning capabilities.
    \item \textbf{Method}: We introduce Self-Backtracking, a novel technique that enables LLMs to internalize the backtracking ability during both training and inference. This approach not only mitigates inefficient overthinking and reduces dependencies on external reward models but also enhances reasoning efficiency by transforming slow-thinking processes into fast-thinking capabilities through self-improvement.
    \item \textbf{Evaluation}: Extensive experimental results on Countdown demonstrate that the proposal can significantly enhance LLMs' reasoning performances, achieving a performance gain of over 40\% compared to the optimal-path SFT method.
\end{itemize}

\section{Related Work}
\paragraph{Learn from Search Trajectories}
Recently, several studies have explored using symbolic search algorithms to construct trajectory data and train transformer models to learn these search strategies, with the aim of enabling models to perform reasoning tasks. For instance, \citet{yang2022chain} employs Monte Carlo Tree Search (MCTS) or BFS to construct reasoning trajectories. Searchformer \cite{searchformer} and DualFormer \cite{dualformer} utilize traces from A* search to train language models, with each trace containing state information, A* heuristic values, and search history. Stream of Search (SoS) \cite{sos} constructs trajectories using various search algorithms to help language models learn the commonalities across different search strategies, facilitating the discovery of new search strategies. GSoS \cite{gsos} further extends SoS by integrating optimal solutions into the process of learning search trajectories.
 DeepSeek R1 \cite{guo2025deepseek} employs end-to-end reinforcement learning to autonomously acquire the capability of search in language. However, training these model to learn exploration trajectories may conflict with guiding it to generate optimal trajectories, leading to inefficient overthinking when solving easy problems.

\paragraph{Learn from Mistakes}
Numerous recent studies have focused on exploring whether language models possess the ability to learn from their previous mistakes and subsequently correct them. One line of techniques \cite{an2023learning,tong2024can,cpo} introduces the external verifier to evaluate the reasoning paths generated by LLMs. This evaluation is then used to construct preference training data for RLHF, with training conducted using algorithms such as PPO \cite{ppo} or DPO \cite{dpo}, enabling self-improvement \cite{yuan2024self} of the models. Another line of techniques \cite{cundy2023sequencematch,zhang2024backtracking,ye2024physics} involves pre-annotating error examples, allowing models to identify whether their current outputs contain issues and adaptively perform backspace during testing to regenerate content. In this paper, our method shows an inherent ability to learn from mistakes and achieves further self-improvement by learning from its search paths.

\paragraph{Inference Strategies of LLM Reasoning}
Many strategies for the inference phase have been proposed to enhance the reasoning capabilities of LLMs. Classical methods such as greedy decoding, beam search \citep{teller2000speech,graves2012sequence}, and majority voting \cite{wang2022selfconsistency} have been widely adopted. Additionally, Best-of-N (BoN) \cite{bonw} is a typical algorithm that generates $N$ complete answers through sampling and selects the optimal one based on the evaluation of a reward model. Recently, approaches that combine search algorithms with LLMs, such as best-first search \cite{tot}, guided beam search \cite{xie2024self} and MCTS \cite{choi2023kcts,feng2023alphazero,zhang2024rest,xie2024monte} have gained increasing attention due to their inference scaling law \cite{wu2024inference,snell2024scaling}. These methods often require additional components such as a verifier, outcome reward model (ORM) \cite{lightman2023let}, or process reward model (PRM) \cite{lightman2023let,wang2024math}, which increases computational cost. This paper proposes a novel method that integrates the verifier within the model to save computational resources while leveraging the advantages of search algorithms, demonstrating scalability in reasoning.

\begin{figure*}[t]
    \centering
    \includegraphics[width=\textwidth]{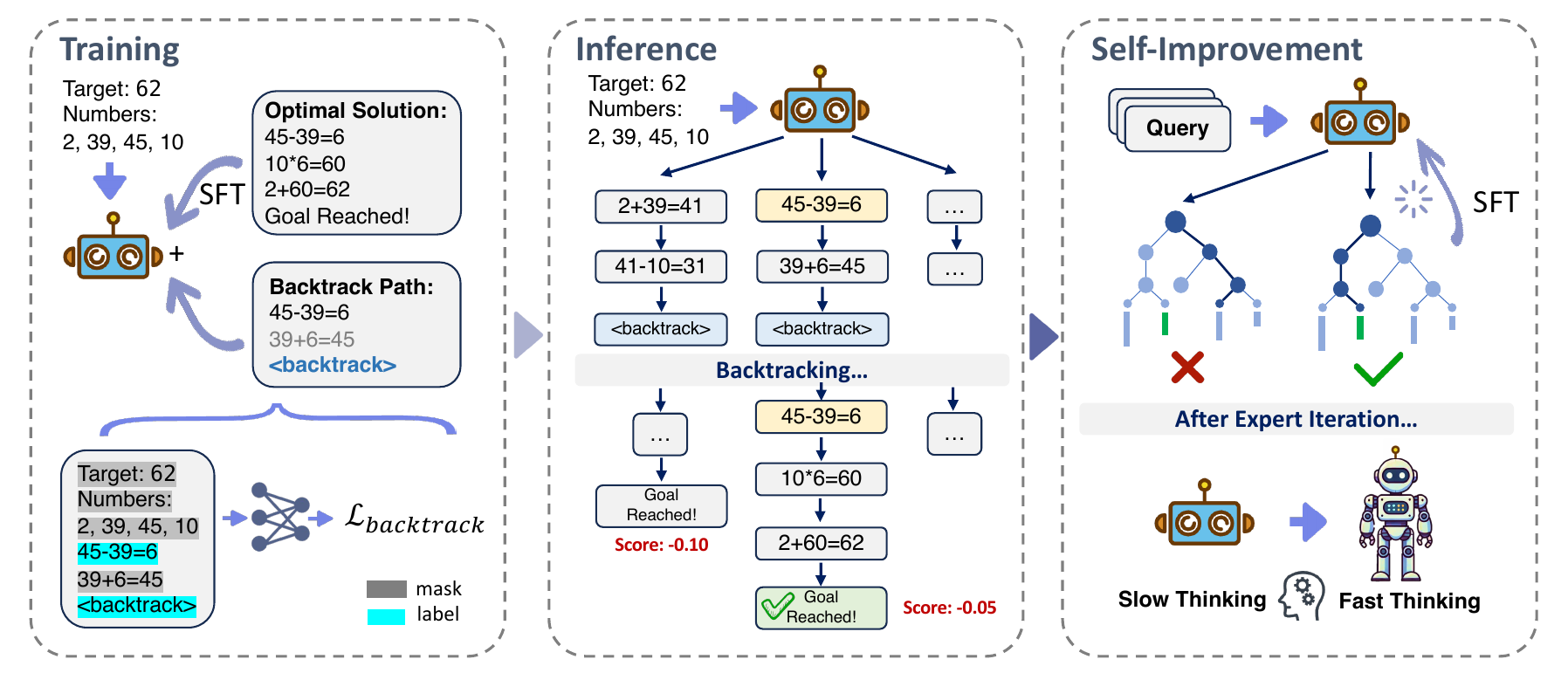}
    \caption{
        \textbf{The overall framework of Self-Backtracking.} During the training phase, the language model is instructed on when and where to backtrack. The inference phase employs a backtracking algorithm that considers both depth and breadth. The self-improvement phase utilizes expert iteration to enhance the fast thinking capabilities of the model.
        }
    \label{method}
\end{figure*}
\section{Preliminary}
\subsection{Problem Setup}
We adopt the formal definition of reasoning problems as proposed in Stream of Search \cite{sos}, where a reasoning problem is modeled as a Markov Decision Process (MDP). An MDP is characterized by the following components: a state set $\mathcal{S}$, representing all possible states within the problem domain; an action set $\mathcal{A}$, denoting all permissible actions; a transition function $T: \mathcal{S} \times \mathcal{A} \rightarrow \mathcal{S}$, which defines how states transition based on actions; and a reward function $R: \mathcal{S} \rightarrow \mathbb{R}$, which assigns a numerical reward to each state.
A typical reasoning task involves an initial state $s_0 \in \mathcal{S}$ and a goal state $s_g \in \mathcal{S}$, where the goal state $s_g$ is associated with a reward of 1 ($R(s_g) = 1$), while all other states yield a reward of zero ($R(s) = 0, \forall s \neq s_g$). The solution $\mathcal P$ to the reasoning problem is represented as a trajectory—a sequence of states and actions—$(s_0,a_0,s_1, a_1, \ldots, s_{g-1},a_{g-1},s_g)$, where each successive state $s_{i+1}$ is determined by the preceding state $s_i$ and a valid action $a_i$ via $s_{i+1} = T(s_i, a_i)$. The trajectory must terminate at the goal state $s_g$.

Let $\Sigma$ represent an alphabet, which is a finite, non-empty set of symbols.  A string is defined as a finite sequence of symbols drawn from $\Sigma$.  In this work, we focus on reasoning tasks that can be expressed purely in the form of language, where both the state set $\mathcal{S}$ and the action set $\mathcal{A}$ consist of strings.  Each state $s \in \mathcal{S}$ could represent an intermediate conclusion or partial solutions and each action $a \in \mathcal{A}$ represents a permissible operation that can be performed on the current state to advance the reasoning process. The transition function is defined as $T(s,a)=s\circ a$, where $\circ$ denotes string concatenation.
\subsection{Backtracking}
Based on the formalization above, we can extend the state-action pairs into a tree, allowing the search process on the tree to be naturally integrated. Backtracking is a classic searching technique that incrementally constructs a solution by exploring various options and retracting decisions when encountering a dead end. This approach is particularly effective in scenarios that require exploring multiple possibilities to solve a reasoning problem, such as navigating a maze or solving puzzles like Sudoku. When the algorithm encounters a dead end, it backtracks to the previous decision point to explore alternative paths, continuing this process until a solution is found or all possibilities are exhausted. Backtracking forms the foundation for many well-known algorithms, including DFS, BFS and MCTS. The general backtracking algorithm is summarized in Algorithm \ref{backtrack_pseudocode}.

\begin{algorithm}[t]
\caption{General Backtracking Algorithm}
\label{backtrack_pseudocode}
\KwIn{Reasoning problem initial state $s_0$}
\KwOut{Solution path $\mathcal{P}$, or \texttt{None} if no solution exists}
\SetKwFunction{FMain}{Backtracking}
\SetKwProg{Fn}{Function}{:}{}
    \Fn{\FMain{$s_0$}}{
    Initialize $\mathcal{P} \gets [s_0]$\;  
    
    \While{True}{
        $s \gets$ last state in $\mathcal{P}$\;  
        
        \If{$s == s_g$}{
            \Return $\mathcal{P}$ \; 
        }
        
        $a \gets$ \texttt{next\_candidate\_action} ($s$)\;  
        
        $s' \gets T(s, a)$\;
    
        \If{\texttt{valid\_state} ($s'$)}{  
                    Append $(a, s')$ to $\mathcal{P}$\; 
        }
        \Else{
            Remove last state from $\mathcal{P}$\;  
            
            \If{$\mathcal{P}$ is empty}{  
                \Return \texttt{None} \; 
            }
            \textbf{continue}\;
        }

    }
}
\end{algorithm}

This study focuses on enabling LLMs to learn backtracking. We identify two core components in general backtracking algorithms: \texttt{next\_candidate\_action} and \texttt{valid\_state}. While the former can be effectively implemented through LLMs' sampling mechanisms, as demonstrated in numerous studies \cite{feng2023alphazero,gptf}, the latter plays a crucial role in evaluating state validity. Traditional search algorithms determine backtracking by identifying terminal states, whereas other approaches, such as A* and MCTS, employ heuristic evaluations to enable early backtracking and accelerate search processes. In this work, we aim to internalize backtracking capabilities within LLMs without relying on external tools (e.g., symbolic verifiers) or models (e.g., reward models) to approximate $\texttt{valid\_state}$. Furthermore, we hope to enhance existing backtracking frameworks through parallel processing to expand search spaces and improve efficiency.

\section{Self-Backtracking in Language Models}
In this section, we introduce our self-backtracking technique. First, during the training phase, we design a specific optimization goal and a tailored dataset format to help the language model learn when and where to backtrack (\textsection \ref{train}) . Second, during the inference phase, we use the learned backtracking ability to create an efficient search algorithm, without the need for additional tools or reward models (\textsection \ref{inference}) . Finally, through a self-improvement process, we feed the search results back into the model, further enhancing its fast-thinking performance (\textsection \ref{improvement}). Figure \ref{method} illustrates the comprehensive framework of our proposed method.
\subsection{Learn to Backtrack}
\label{train}
Under the standard Supervised Fine-Tuning (SFT) framework, we typically employ a dataset $\mathcal{D}_{op} = \{(x_i, y_i)\}_{i \in [n_{op}]}$, where for reasoning tasks, $y_i$ represents the natural language reasoning path representing the optimal solution. To enable the model to backtrack at appropriate times and positions, we introduce a backtracking  dataset: 
$$\mathcal{D}_{back} = \{\left(x_j, \texttt{prefix}(y_j) \circ a_{err} \circ \langle \texttt{backtrack} \rangle\right)\}_{j \in [n]}$$
Here, $\texttt{prefix}(y_j)$ denotes the prefix of the optimal solution $y_j$, representing a partial solution; $a_{err}$ signifies an erroneous action extended from the partial solution, which cannot lead to the correct answer; and $\langle \texttt{backtrack} \rangle$ is a special token indicating that the model needs to backtrack from the current state. The final dataset is $\mathcal {D}=\mathcal D_{op}\cup \mathcal D_{back}$.
For different tasks, there are various methods for constructing $\mathcal{D}_{back}$.
In our experimental setup, the questions for both $\mathcal{D}_{op}$ and $\mathcal{D}_{back}$ are configured to be identical. This configuration allows us to effectively model the dataset as a preference dataset, so we can compare more baselines.

Through this data construction, if the model learns to recognize $a_{err}$ and utilize the $\langle \texttt{backtrack} \rangle$ token for backtracking, it acquires the ability of when to backtrack, fulfilling the requirement of \texttt{valid\_state}. Simultaneously, this dataset format implicitly contains information on where to backtrack, indicating that the model should revert to the state represented by $\texttt{prefix}(y_j)$. Although this design superficially appears to support only single-step backtracking, the recursive nature of the backtracking algorithm allows the model to achieve multi-step backtracking once it masters single-step backtracking. 


For the given dataset $\mathcal{D}$, we formulate the training loss function for the language model parameterized by $\theta$ as follows:
\begin{equation}
\mathcal{L}(\theta) = \mathcal{L}_{SFT}(\theta)+\mathcal{L}_{backtrack}(\theta)
\end{equation}
The loss function comprises two primary components: firstly, the SFT loss: $$\mathcal{L}_{SFT}=-\frac{1}{n_{op}}\sum_{i=1}^{n_{op}}\log p_{\theta}(y_i|x_i),$$
which aims to encourage the model to generate corresponding reasoning steps and final answers based on given questions. The second loss term $\mathcal{L}_{backtrack}$ contains two parts:

\scalebox{0.91}{%
$\begin{aligned}
&\mathcal{L}_{backtrack}(\theta) = -\frac{1}{n_{back}}\sum_{j=1}^{n_{back}}\log p_{\theta}(\texttt{prefix}(y_j)|x_j)\\ &
-\frac{1}{n_{back}}\sum_{j=1}^{n_{back}} \log p_{\theta}(\langle \texttt{backtrack} \rangle | x_j \circ \texttt{prefix}(y_j) \circ a_{err})
\end{aligned}$%
}
 
The first part targets partially correct reasoning paths in backtracking samples, designed to enable the model to accurately predict partial solutions given the input. The second part focuses on the model's ability to predict the $\langle \texttt{backtrack} \rangle$ token when it has deviated from the correct path, encouraging the model to learn when and where to backtrack.  Notably, compared to the SFT loss applied on the $\mathcal{D}_{back}$ dataset, the combination of the second and third loss terms omits the loss component for predicting $a_{err}$.  This design is reasonable as our objective is not to encourage the model to predict incorrect actions but to prevent it from falling into erroneous reasoning paths.  In practical implementation, this can be achieved by applying a mask to $a_{err}$ when computing SFT loss, as illustrated in Figure \ref{method}.

\subsection{Inference with Backtracking}
\label{inference} 
Upon learning when and where to backtrack, the backtracking algorithm (see Algorithm \ref{backtrack_pseudocode}) can be integrated into the inference search process. We further propose a self-backtracking inference algorithm that consider both depth and breadth search, which primarily consists of three steps: \textbf{Expansion}, \textbf{Backtracking}, and \textbf{Selection}.

\paragraph{Expansion.}
In the expansion phase, given the current state \( s \), the algorithm samples \( N \) predictions from the language model. These predictions are then categorized into two groups: those containing the \(\langle \texttt{backtrack} \rangle\) token and those that do not. Predictions without the \(\langle \texttt{backtrack} \rangle\) token are directly added to the candidate set, while those containing the token are processed further in the next phase.

\paragraph{Backtracking.}
During the backtracking phase, the algorithm selects \(\sqrt{N}\) predictions containing the \(\langle \texttt{backtrack} \rangle\) token. These predictions are rolled back by one reasoning step and re-expanded, returning to the expansion phase. This iterative process is repeated \( b \) (predefined budget) times to expand the candidate set.

\paragraph{Selection.}
Finally, in the selection phase, we compute the scores for all candidate reasoning paths by utilizing the negative perplexity as the metric, and subsequently return the result with the highest score.

This algorithm enables a flexible search mechanism, where the breadth of exploration is governed by parameter $N$ and the depth by parameter $b$. It leverages the inherent backtracking capabilities learned by the language model during training without requiring an external reward model, while maintaining controllable computational costs throughout the generation process.
\subsection{Self-Improvement}
\label{improvement}
In this stage we aim to transfer the model's slow thinking abilities to fast thinking through the self-improvement method. To achieve this, we employ an expert iteration strategy, which primarily consists of three steps: First, during the slow thinking data generation phase, we utilize the self-backtracking inference model to produce high-quality reasoning path data. Subsequently, in the expert screening phase, experts evaluate the generated data to select training samples suitable for the fast thinking model. In our experiment, we quantify the model's accuracy using an evaluator. Finally, in the fast thinking model training phase, the selected high-quality data is used to train the fast thinking model by SFT. Through this iterative optimization, we get continuous enhancement in the performance of the fast thinking model. The process is shown in Algorithm \ref{alg:ei}.

\begin{algorithm}[t]
\caption{Expert Iteration for Self-Improvement}
\label{alg:ei}
\KwIn{Self-backtracking model $M_0$, initial Dataset $\mathcal{D}_0$, number of iterations $K$}
\For{$t \leftarrow 0$ \KwTo $K-1$}{$\mathcal{D}_t \leftarrow \{M_t(x_i) \mid x_i \in \mathcal{D}_0\}$\\$\tilde{D}_t \leftarrow \{(x_i, p_i) \mid (x_i, p_i) \in \mathcal{D}_t$, path $p_i$ is correct $\}$\\
$M_{t+1} \leftarrow \text{SFT}(M_t, \tilde{D}_t)$
}
\KwOut{Optimized fast thinking model $M_K$}
\end{algorithm}

\section{Experiments}
\label{exp}
\begin{table*}[t]
\centering
\caption{\textbf{Self-Backtracking Enhances Reasoning Performance.} We report the accuracy (\%) for the countdown task across two base models (Llama3.2-1B and Llama3.2-3B) with several baseline models. Best results for each base model are \textbf{bolded}.}
\label{maintab}
\begin{tabular}{lcccc}
\toprule
\multirow{2}{*}{\textbf{Methods}} & \multicolumn{2}{c}{\textbf{Llama3.2-1B}}& \multicolumn{2}{c}{\textbf{Llama3.2-3B}}\\
\cmidrule{2-5}
& {\textbf{Seen Targets}} & {\textbf{New Targets}} & {\textbf{Seen Targets}} & {\textbf{New Targets}} \\
\midrule
\multicolumn{5}{l}{\(\triangleright\) \textit{Only Optimal Solution:}} \\
\midrule
  SFT + Greedy & 28.60 & 28.92 & 33.98 & 32.68 \\
  SFT + Beam Search & 31.68 & 31.90 & 35.82 & 34.36 \\
\midrule
  \multicolumn{5}{l}{\(\triangleright\) \textit{RLHF}} \\
\midrule
  DPO \cite{dpo}& 29.06 & 27.64
  & 34.46 & 32.72 \\
KTO \cite{kto} &  28.34 & 27.74 & 33.70 & 32.34\\
 Best-of-N ($N=8$)& 41.26 &40.68  & 47.84 &48.56  \\
 Best-of-N ($N=16$)&32.40 & 33.94&47.28  & 47.80 \\
  Best-of-N ($N=32$)&25.60 &27.04 &44.38  &45.88  \\
\midrule
  \multicolumn{5}{l}{\(\triangleright\) \textit{Backtracking Data:}} \\
\midrule
Greedy & 28.92 & 27.06 &27.56  & 26.96 \\
Beam Search & 36.10 & 34.30 & 23.10 & 22.20 \\
Self-Backtracking ($b=0,N=8$) & 66.66 &       67.40 &58.76 & 56.92 \\
Self-Backtracking ($b=0,N=32$) & 70.66 &    72.14 & 60.18& 58.06  \\
Self-Backtracking ($b=1,N=8$) & 67.60 &       68.02 &61.06 & 59.28 \\
Self-Backtracking ($b=1,N=32$) & \textbf{73.54} &  \textbf{73.78} & \textbf{64.12} & \textbf{61.98} \\
\bottomrule
\end{tabular}

\vspace{-10pt}
\end{table*}
In this section, we evaluate the capability of the self-backtracking algorithm to enhance the reasoning performance of language models on the Countdown \cite{sos,gsos} task. This task requires the language model to exhibit robust reasoning abilities, also posing a substantial challenge even for humans. We also analyze that our method exhibits the test-time scaling law and possesses the ability to self-improvement, demonstrating significant advantages over other approaches.
\subsection{Experimental Setup}
\paragraph{Dataset.} We employ the Countdown task \cite{sos} as our principal experimental framework to rigorously evaluate the reasoning capabilities of our self-backtracking approach. This task extends the traditional 24 Game \cite{yang2022chain} by necessitating that LLMs strategically combine a provided set of input numbers using fundamental arithmetic operations—addition, subtraction, multiplication, and division—to achieve a predefined target number. The complexity of the task stems from its expansive search space, which rigorously tests the models' ability in reasoning the correct path. We construct datasets focusing on problem instances with four input numbers and target values $\leq$ 100. The training set consisted of 500,000 samples, balanced between optimal solutions and backtracking data. The test set was systematically partitioned into two distinct categories: one comprising seen targets paired with novel input combinations (denoted as \textbf{Seen Targets}), and the other incorporating entirely new targets (denoted as \textbf{New Targets}), consistent with the setting outlined in SoS \cite{sos}. Each category contained 5,000 instances.

\paragraph{Data Construction.}
In the construction of optimal solutions, we first randomly generate the target value and then select four operands within a possible range. To solve the mathematical problem, we use a recursive exhaustive method to systematically explore all potential solutions.
The construction of backtracking data is categorized into three types based on different types of error patterns:
1) Exploration Errors: For a given set of target numbers and operands, we employ a DFS strategy to generate search steps. If the generated steps do not match the correct solution steps, subsequent searches are terminated.
2) Computational Errors: These errors are constructed by inserting invalid mathematical equations within the solution steps.
3) Rule Violations: This type of error is created by deliberately using operands not in the predefined list of available operands in the solution steps, thereby violating the solution rules.
After appending $\langle \texttt{backtrack} \rangle$ tokens immediately following the erroneous steps in all backtracking samples, the final training dataset is formed, with the respective proportions of the above error modes being 1:2:2.

\paragraph{Comparison Methods}
In the countdown task, we employ Llama3.2-1B \cite{llama3} and Llama3.2-3B \cite{llama3} as the base models. The comparative experiments primarily consist of three categories of methods: The first category involves supervised fine-tuning (SFT) using only optimal solution dataset $\mathcal{D}_{op}$, and compares two typical sampling strategies, namely greedy search and beam search (beams=16). The second category models the data as preference data pairs and compares various RLHF algorithms, including DPO \cite{dpo}, KTO \cite{kto}, and the Best-of-N \cite{bonw} selection method based on the outcome reward model. The third category is based on backtracking data and compares the greedy sampling strategy. Furthermore, we compare the symbolic solver DFS under specified budgets (backtrack limits $b=32$ and $b=64$) with search-augmented methods, SoS \cite{sos} and GSoS \cite{gsos}, using their reported optimal performances.

\paragraph{Experimental Details}
Our self-backtracking algorithm is implemented using PyTorch with Deepspeed Stage 2 optimization. Training is conducted on four NVIDIA A800 GPUs, using a base learning rate of 1e-5 over three epochs. Model-specific precision is applied: FP32 for Lama3.2-1B and BF16 for Lama3.2-3B. Input sequences are truncated to 128 tokens. Detailed training specifications and baseline implementations are provided in the Appendix \ref{app:train}. During the inference phase, we employ beam search with a temperature of 0.7 for our method and baselines involving sampling.
For an analysis of temperature stability, additional experiments are provided in Appendix \ref{app:temp}.
\begin{table}[t]
\centering
\caption{Comparison between search-augmented methods.}
\label{small}
\resizebox{\columnwidth}{!}{ 
\begin{tabular}{lcc}
\toprule
\textbf{Methods} & \textbf{Seen Targets} & \textbf{New Targets} \\
\midrule
DFS ($b=32$) & 49.12 & 48.90 \\
DFS ($b=64$) & 60.00 & 61.06 \\
SoS \cite{sos} & 57.50 & 53.40 \\
GSoS \cite{gsos} & 69.00 & 67.20 \\
Self-Backtracking (best) & \textbf{73.54} & \textbf{73.78} \\
\bottomrule
\end{tabular}}
\end{table}
\subsection{Main Results}
\begin{figure}[t]
    \centering
\includegraphics[width=0.50\textwidth]{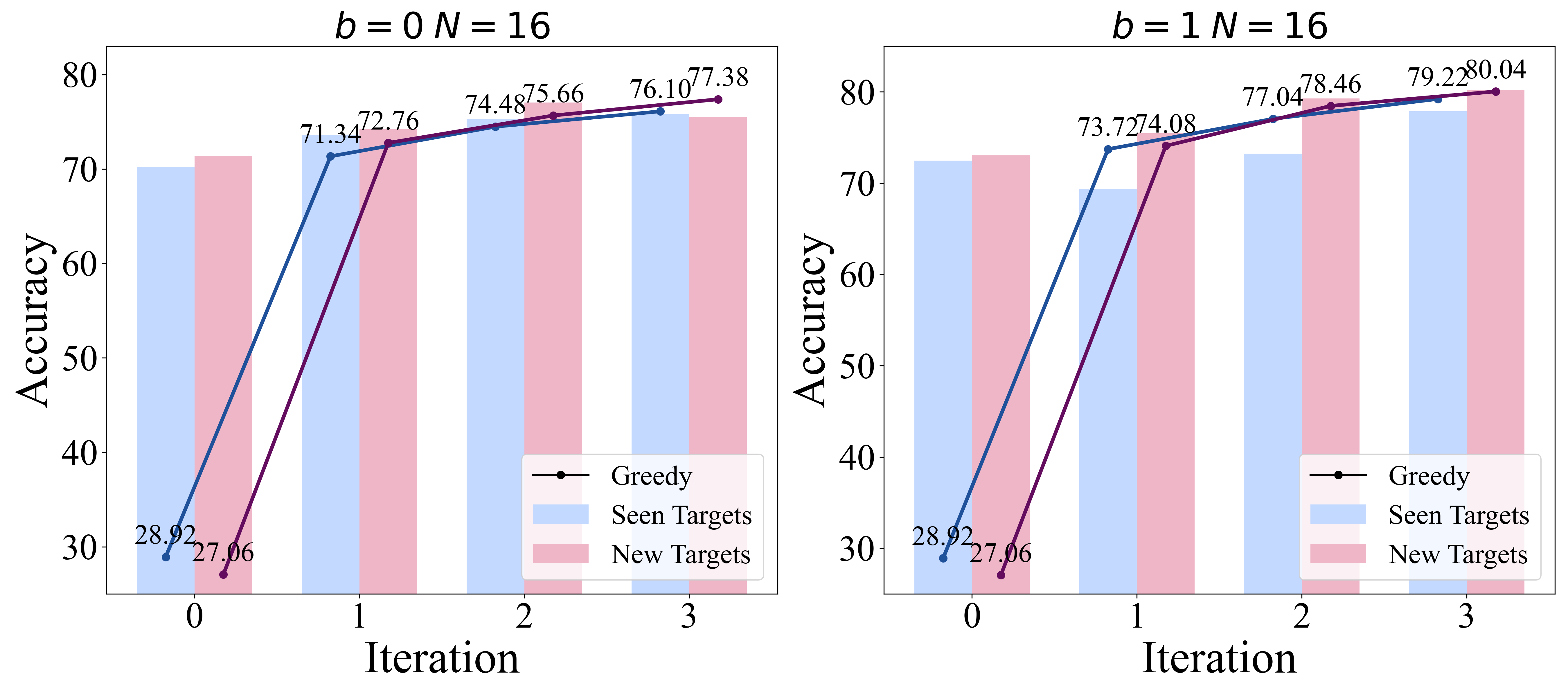}
\vspace{-10pt}
    \caption{Self-improvement in accuracy of Llama3.2-1B.}
    \label{fig:improvement}
\end{figure}
\begin{figure}[htbp]
    \centering
\includegraphics[width=0.485\textwidth]{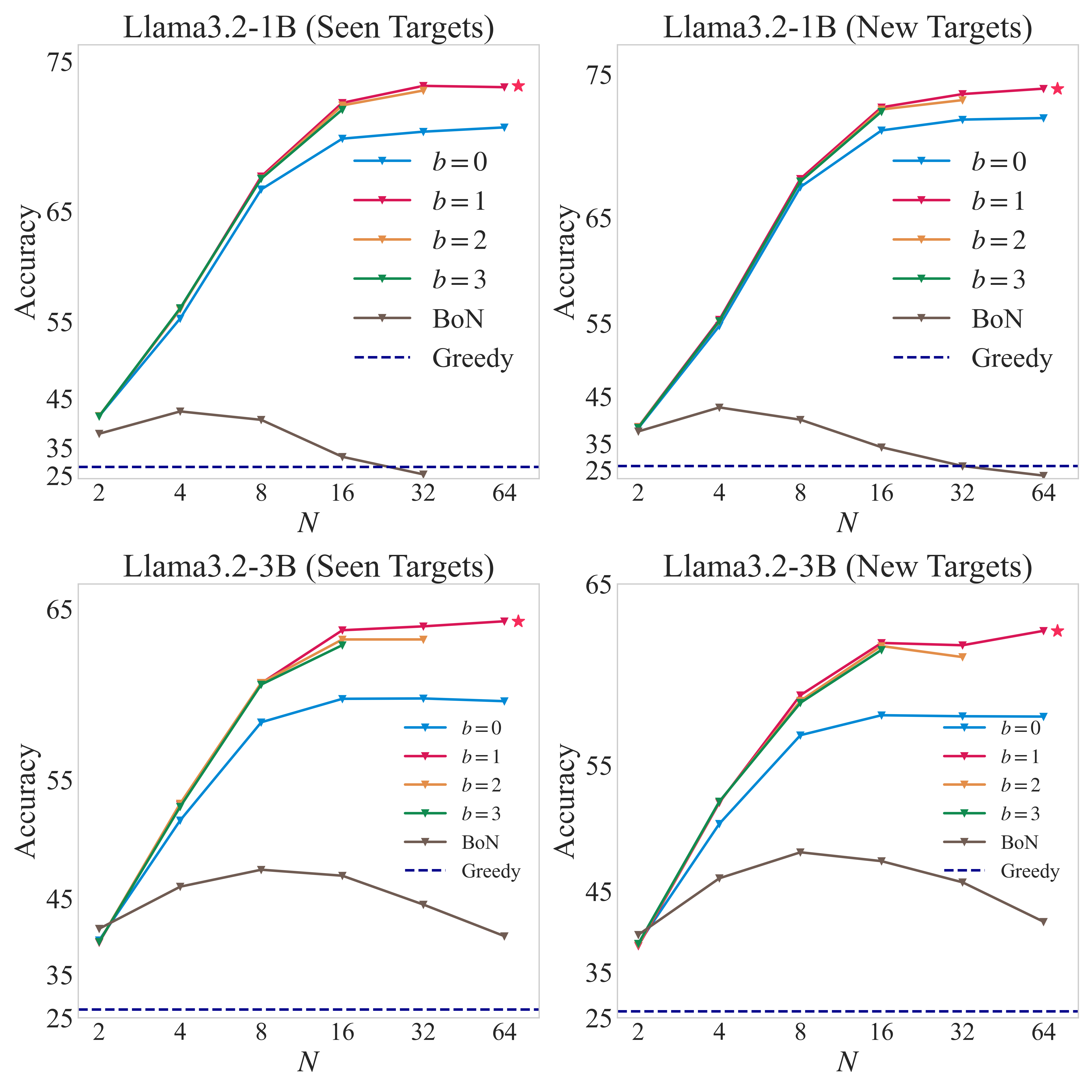}
\vspace{-10pt}
    \caption{The performance curves when varing $N$.}
    \label{fig:bn}
\end{figure}
\subsubsection{Self-Backtracking Boosts Reasoning}
In Table \ref{maintab}, we present the accuracy of various methods across different models for the countdown task.
Overall, the self-backtracking technique demonstrates a significant improvement over the baseline of greedy search after SFT, with enhancements of approximately 40\% on Llama3.2-1B and over 30\% on Llama3.2-3B. Notably, our method exhibits considerable advantages even when \( b = 0 \), i.e., without backtracking, suggesting that the \texttt{<backtrack>} token can implicitly assess the quality of the current state, effectively substituting the function of the reward model. Additionally, when \( b = 1 \), we find that performance further improves, indicating that backtracking to previous states enables the model to leverage search mechanisms to explore correct answers more effectively. 
The experimental results yield a strange finding: our algorithm's Llam3.2-3B underperforms compared to Llam3.2-1B, although it is larger. We observe that while the 3B model demonstrates superior computational accuracy, it always fails to achieve the target values. This phenomenon, which will be further examined in subsequent sections through the error type analysis.

In addition, we compared the best performance of our method (Llama3.2-1B, $b=1, N=32$) with the classical symbolic algorithm BFS (with a backtracking budget) and search-augmented methods, as shown in Table \ref{small}. The results demonstrate that our approach exhibits significant advantages over these algorithms in such reasoning task, even those already utilizing search frameworks.

\subsubsection{Self-Backtracking can Self-Improve}
Further experiments demonstrate that our algorithm achieves self-improvement through expert iteration. Employing self-backtracking with configurations $b=0, N=16$ and $b=1, N=16$ on two base models and datasets respectively, we filtered correct reasoning paths from inference outputs for SFT. Figure \ref{fig:improvement} presents three-round improvement results, where bars indicate test performance using our algorithm (slow thinking) and lines represent greedy search (fast thinking). Each iteration brings moderate gains (1-2\%) for slow thinking, while fast thinking achieves remarkable improvements: +40\% after first iteration (approaching slow thinking's performance), ultimately surpassing slow thinking by 50\% relative gain after third iteration. Notably, $b=1$ significantly outperforms $b=0$, confirming backtracking's importance. This evidences our method's capability to distill slow-thinking advantages into fast-thinking models through expert iteration, enabling optimal single-pass inference.

\begin{figure}[t]
  \centering
\includegraphics[width=0.485\textwidth]{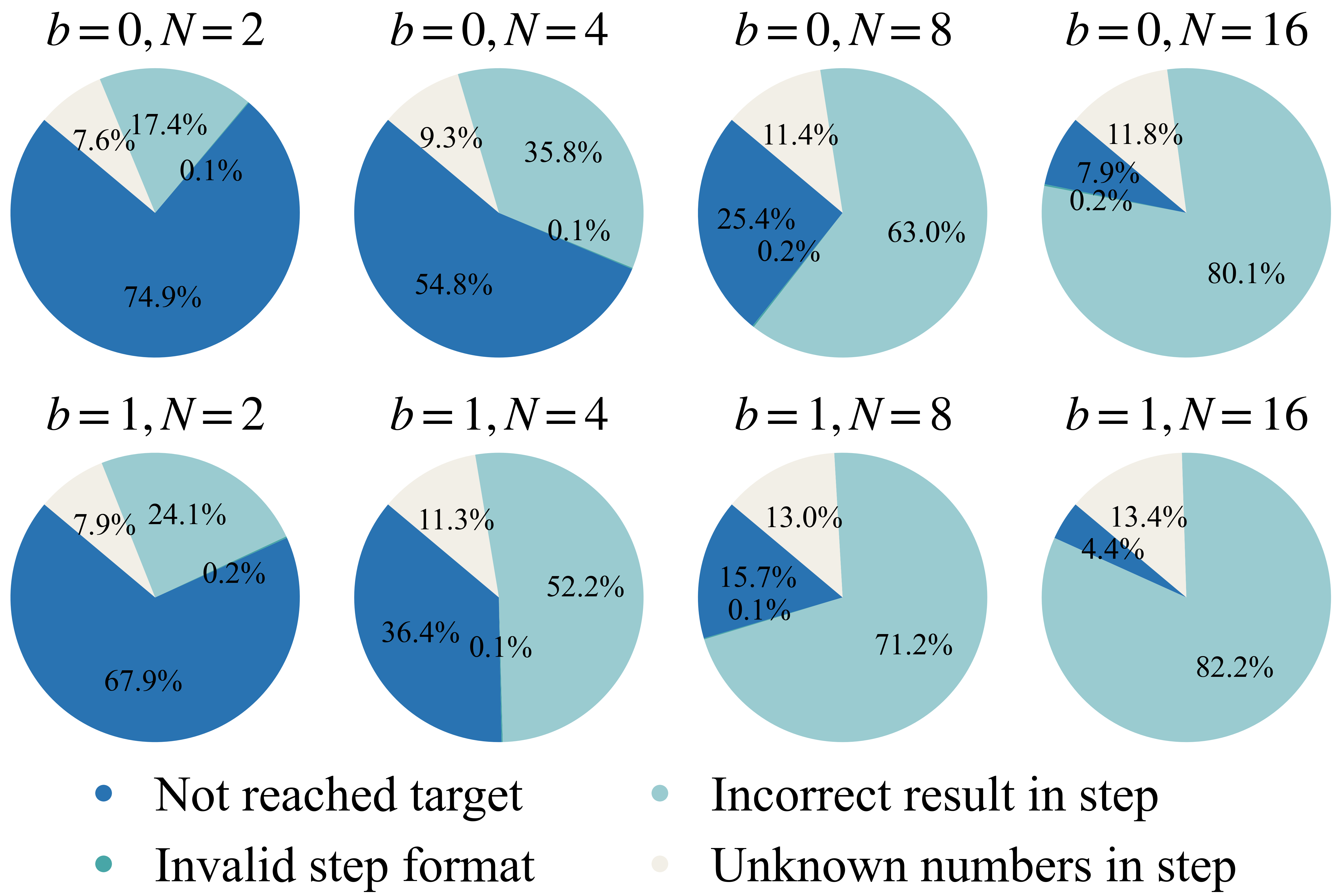}
\vspace{-10pt}
  \caption{Error types of our method for different $b$ and $N$ on Seen Targets of Llama3.2-1B.}
  \vspace{-10pt}
  \label{fig:pie}
\end{figure}
\subsection{Analysis}
\paragraph{Analysis for different $b$ and $N$.}
We conduct experiments by varying the $b$ and $N$, and generate performance curves under different $b$ values as $N$ increases, as illustrated in Figure \ref{fig:bn}. The results demonstrate that the performance of BoN initially increases and then decreases with larger $N$, which we attribute to the reward hacking. On the contrary, our method exhibits a consistent improvement with increasing $N$, eventually stabilizing, indicating a clear test-time scaling law in breadth. Furthermore, when backtracking is permitted ($b=1$), the performance improves more rapidly with $N$ and achieves a higher overall performance, underscoring the necessity of backtracking. Surprisingly, increasing $b$ does not result in a significant scaling phenomenon in depth. Our case study reveals that the diversity of outputs from secondary backtracking significantly decreases, leading to only marginal improvements compared to $b=1$.

\paragraph{Analysis for error types.}

We conduct experiments to analyze the error types for different $b$ and $N$. There are four error types: not reached target, invalid step format, incorrect result in step and unknown numbers in step.
The experimental results shown in Figure \ref{fig:pie} demonstrate that our method progressively reduces the proportion of ``Not reached target" errors by expanding the search scope. As the parameter $N$ increases, the model explores more nodes. Allowing more backtracking provides the model with additional opportunities for error correction and retreat, thereby enhancing the probability of finding the correct answer. However, there is an increase in the proportion of computational errors, stemming from the model's improper alteration of the final step results in order to approximate the target value. In contrast, the proportions of errors involving the use of non-compliant operands and formatting errors remain low and stable, having a limited impact on overall performance. The whole results can be seen in Appendix \ref{app:error}.

\begin{table}[t]
\centering
\caption{Performance evaluation of self-backtraking under varying ratios of $\mathcal{D}_{op}$ to $\mathcal{D}_{back}$ for Llama3.2-1B on Seen Targets.}

\label{neg}
\resizebox{\columnwidth}{!}{ 
\begin{tabular}{lccccc}
\toprule
\textbf{Parameter} & \textbf{1 : 0.5} & \textbf{1 : 1} & \textbf{1 : 2} & \textbf{1 : 3} & \textbf{1 : 4} \\
\midrule
$b=0, N=8$        & \textbf{66.70}  & 66.66  & 64.92  & 64.78  & 65.06  \\
$b=1, N=8$        & \textbf{68.00}  & 67.60  & 64.66  & 65.18  & 65.94  \\
$b=0, N=16$       & 69.80  & \textbf{70.20}  & 67.80  & 68.18  & 69.16  \\
$b=1, N=16$       & 71.12  & \textbf{72.50}  & 65.66  & 66.56  & 70.12  \\
\bottomrule
\end{tabular}}
\vspace{-10pt}
\end{table}
\paragraph{Analysis for different ratio between $\mathcal{D}_{op}$ and $\mathcal{D}_{back}$.}

To investigate the impact of backtracking data, we conducted additional experiments by controlling the quantity of $\mathcal{D}_{back}$. We examined five ratios of $\mathcal{D}_{op}$ to $\mathcal{D}_{back}$: 1:0.5, 1:1, 1:2, 1:3, and 1:4, testing them on Llama3.2-1B while keeping all training parameters consistent. The experimental results, as shown in \ref{neg}, indicate that while an increase in backtracking data slightly diminishes the performance of our algorithm, the overall impact is minimal. Notably, the 1:0.5 ratio yields the best results when $N=8$, whereas the 1:1 ratio is optimal when $N=16$. We attribute this to the fact that with more backtracking data, the model can achieve more accurate results, thus benefiting from increased sampling. In conclusion, taking into account both the effectiveness and computational cost of training, we recommend maintaining a $\mathcal{D}_{op}$ to $\mathcal{D}_{back}$ ratio of no less than 1.

\section{Conclusion}
In this study, we propose a novel Self-Backtracking technique that addresses critical limitations in current reasoning models by enabling them to internalize the search process, particularly the ability to autonomously determine when and where to backtrack. This approach not only mitigates inefficient overthinking and reduces dependencies on external reward models but also enhances reasoning efficiency by transforming slow-thinking processes into fast-thinking capabilities through self-improvement. Empirical evaluations on the Countdown task demonstrate that our method achieves a performance gain of over 40\% compared to the optimal-path supervised fine-tuning baseline, highlighting its effectiveness in improving reasoning capability.

\paragraph{Limitations and future work.}
This study has several limitations. For instance, the method has not been adapted to a broader range of general reasoning tasks, and need further scaling up. We plan to demonstrate the advantages of our method in more general reasoning tasks on larger LLMs in subsequent research.

\section{Impact Statements}
This work advances LLMs reasoning by enabling the model to backtrack during both the training and inference phases. Our method has the potential to enhance a wide range of applications that require reliable artificial intelligence reasoning, such as mathematical problem-solving. There are many potential societal consequences of our work, none which we feel must be specifically highlighted here.
\bibliography{example_paper}
\bibliographystyle{icml2025}

\appendix
\onecolumn
\section{Training Details}
\label{app:train}
 All baseline methods are trained on a cluster with four NVIDIA A800 GPUs. Model-specific precision is applied: FP32 for Lama3.2-1B and BF16 for Lama3.2-3B. Input sequences are truncated to 128 tokens.
\begin{itemize}
    \item \textbf{Hyper-parameters:}
    \begin{itemize}
        \item Learning rate: $1 \times 10^{-5}$
        \item Warmup steps: 1
        \item Batch size: 16
        \item Learning rate scheduler: Cosine
        \item Training epochs: 3
    \end{itemize}
    \item \textbf{For DPO:} $\beta = 0.5$, RPO $\alpha= 1.0$.
\end{itemize}

\begin{figure}[htbp]
  \centering
\includegraphics[width=0.8\textwidth]{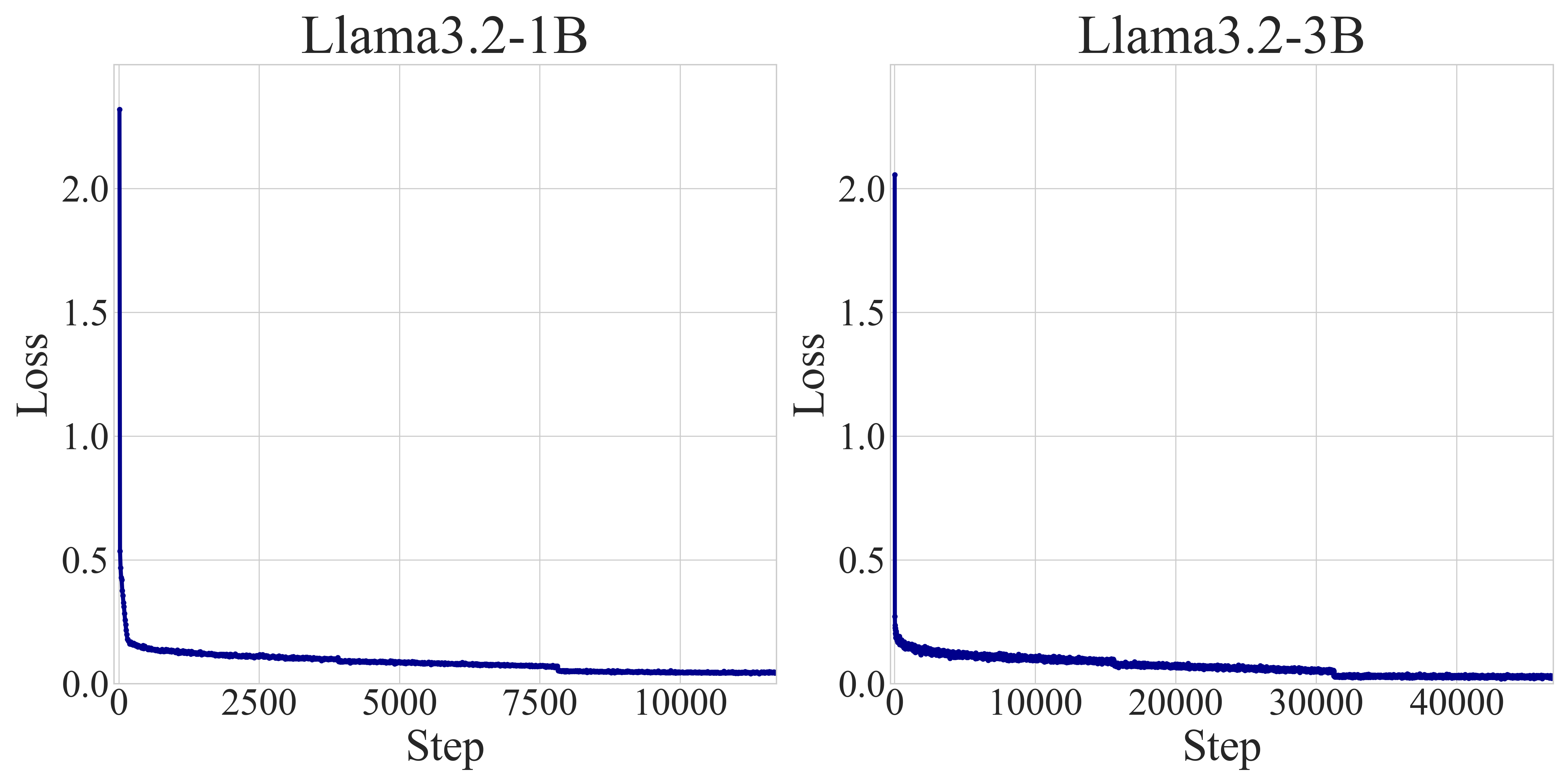}
  \caption{Loss Curve of self-backtracking for Llama3.2-1B and Llama3.2-3B.}
  \label{fig:loss}
\end{figure}

We also demonstrate the loss curve in Figure \ref{fig:loss} of the training process of our self-backtracking method.

\section{Case Study}
\begin{figure}[htbp]
    \centering
\includegraphics[width=0.65\textwidth]{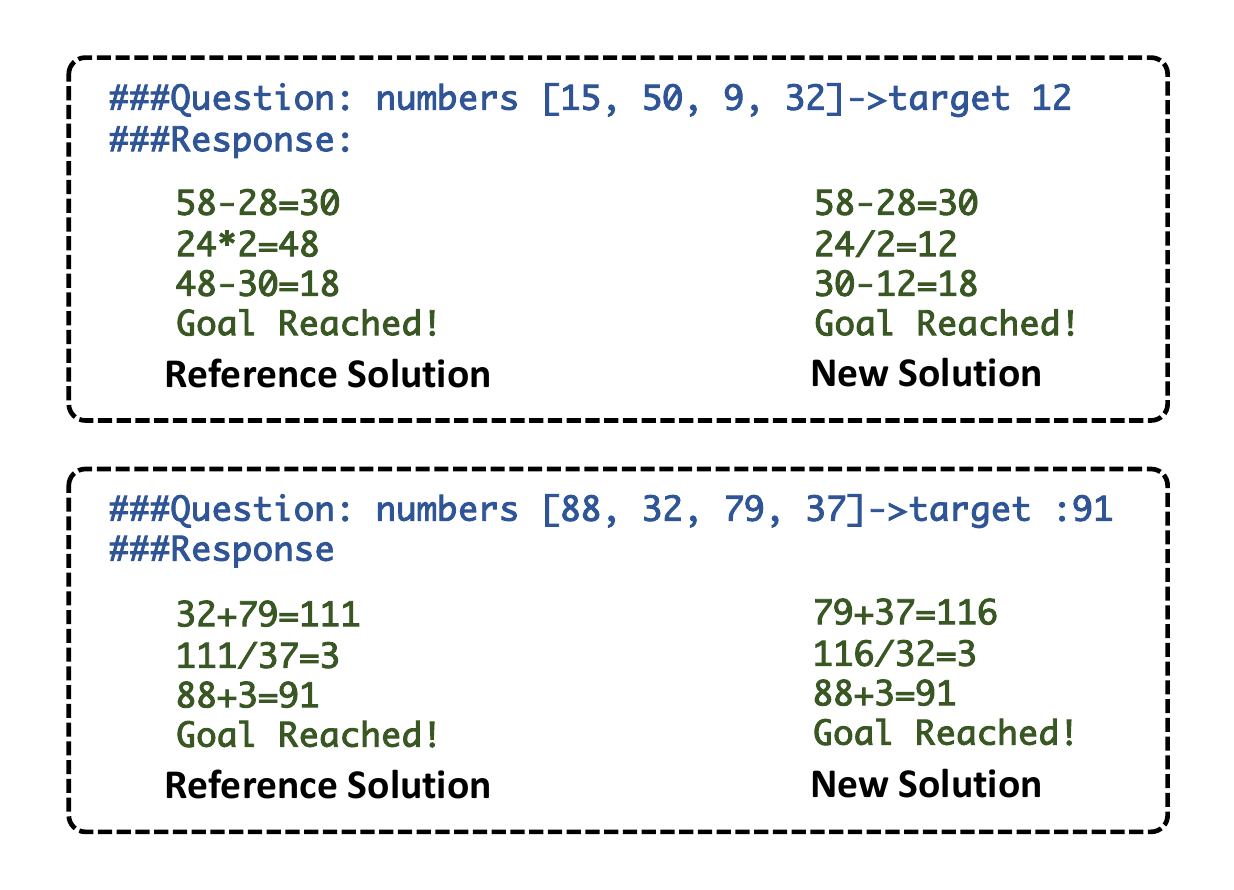}
    \caption{Case Study of our self-backtracking.}
    \label{fig:case}
\end{figure}

We present two illustrative examples in Figure \ref{fig:case}, both derived from our self-backtracking approach. Our analysis reveals that the proposed model is capable of identifying novel solutions distinct from the reference solutions, thereby demonstrating the superior reasoning capabilities of our algorithm.

\label{app:case}
\section{More Results}
\subsection{Self-Improvement}
\label{app:im}
\begin{figure}[htbp]
    \centering
\includegraphics[width=0.80\textwidth]{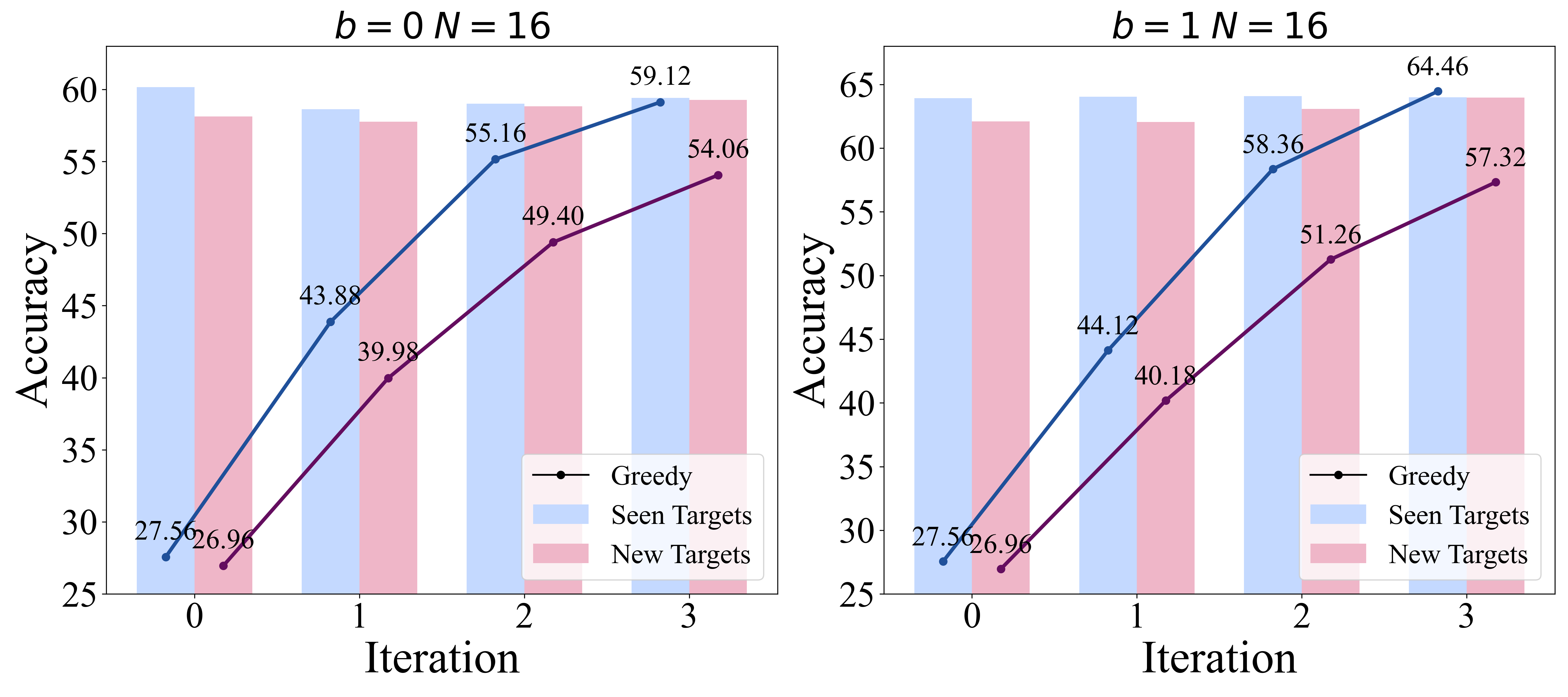}
    \caption{Self-improvement in accuracy of Llama3.2-3B.}
    \label{fig:improvement2}
\end{figure}

We supplement the experimental results of expert iteration on Llama3.2-3B in Figure \ref{fig:improvement2}.
\subsection{Analysis for Error Types}
\label{app:error}
\begin{figure}[htbp]
  \centering
\includegraphics[width=0.9\textwidth]{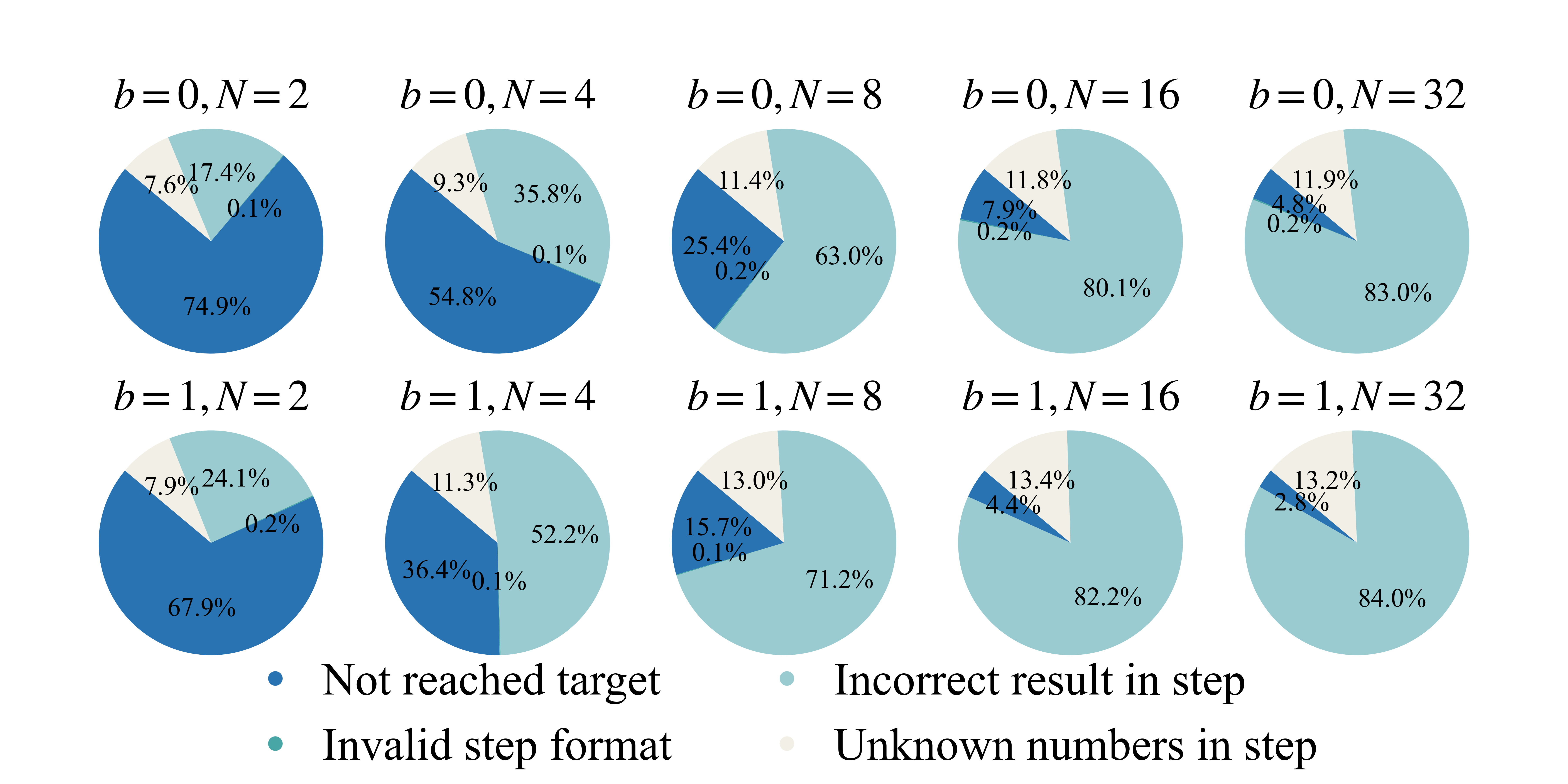}
  \caption{Error types of Llama3.2-1B on seen targets.}
  \label{fig:1seen}
\end{figure}

\begin{figure}[htbp]
  \centering
\includegraphics[width=0.9\textwidth]{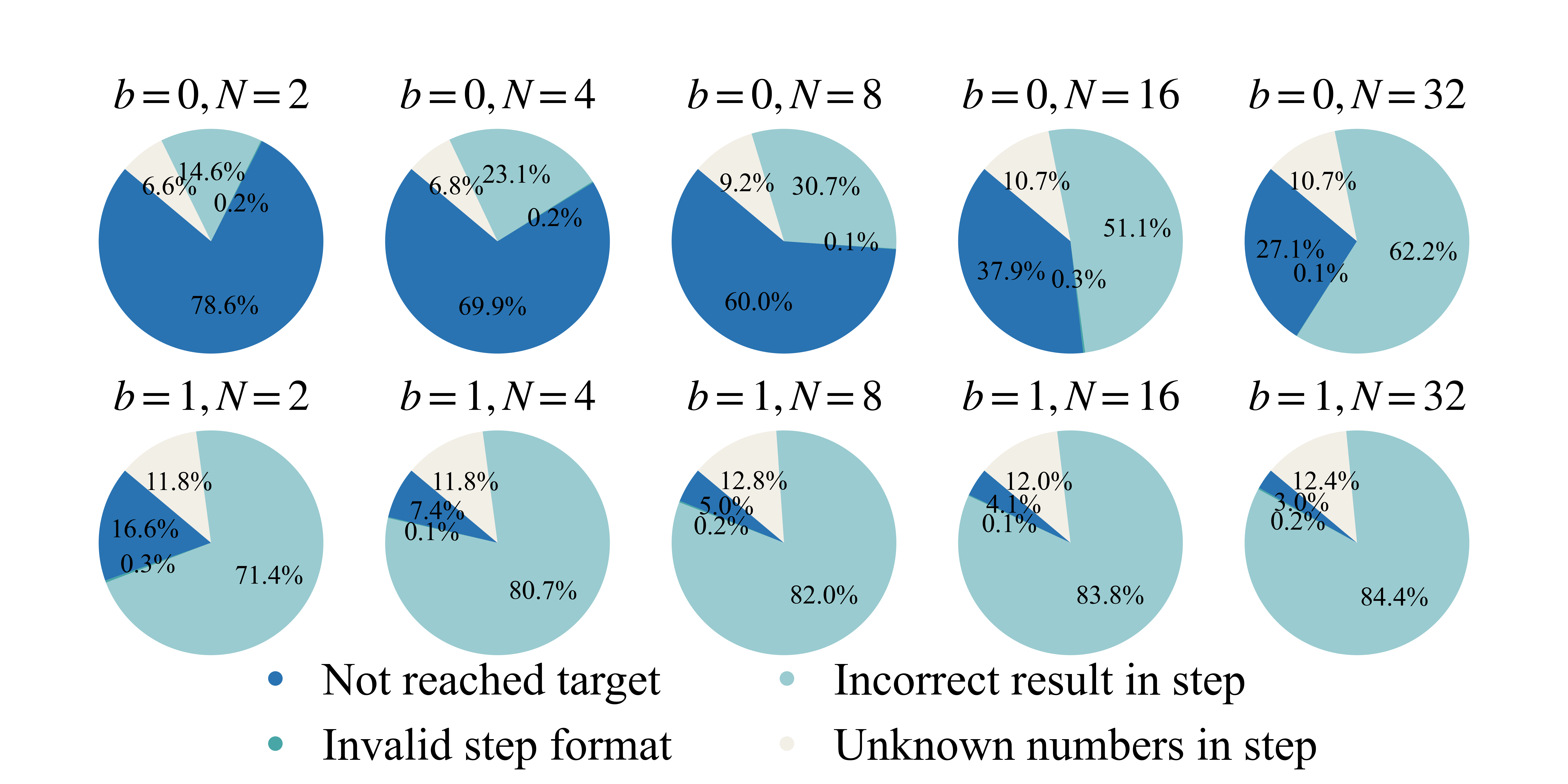}
  \caption{Error types of Llama3.2-1B on new targets.}
  \label{fig:1new}
\end{figure}

\begin{figure}[htbp]
  \centering
\includegraphics[width=0.9\textwidth]{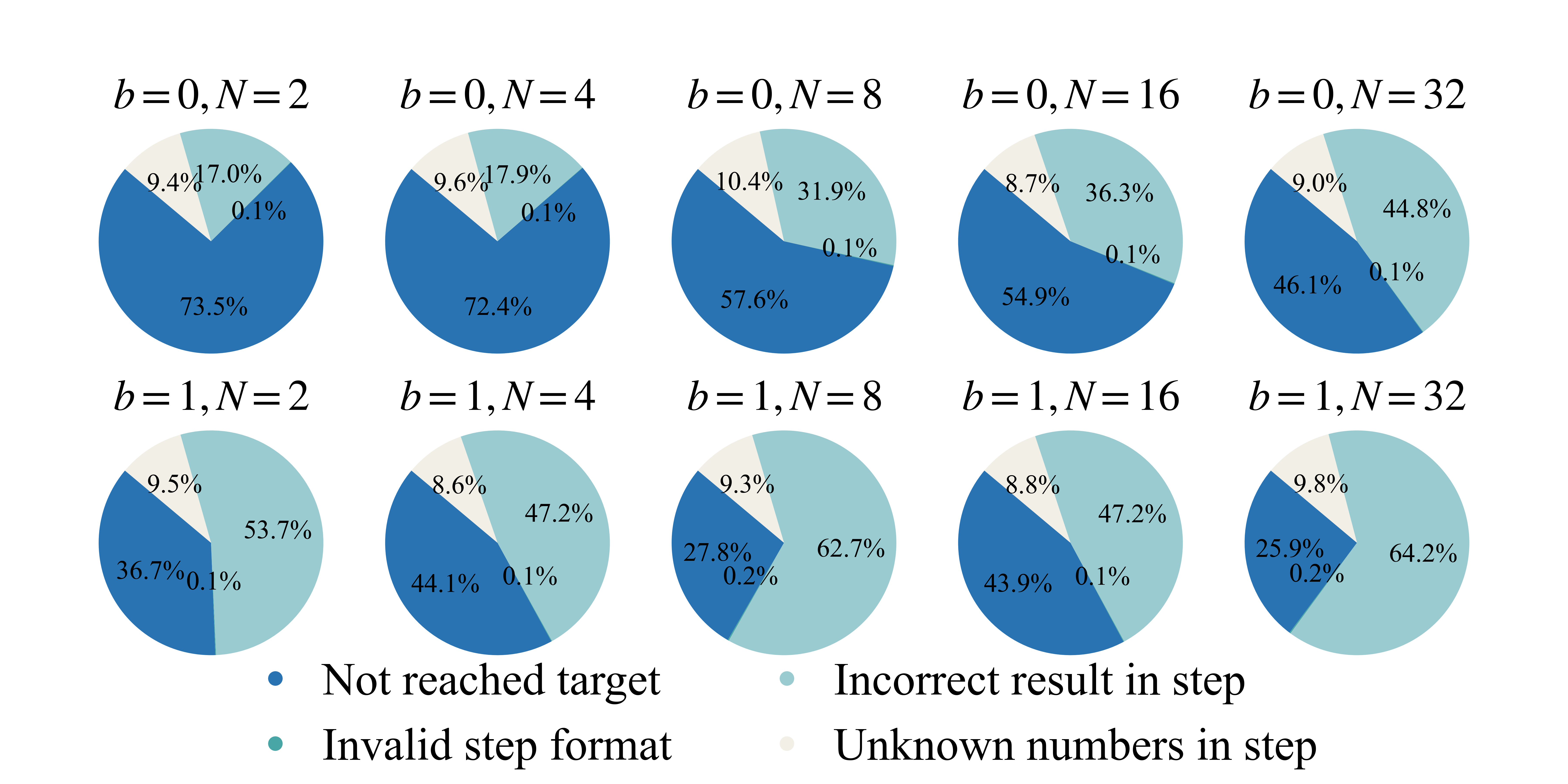}
  \caption{Error types of Llama3.2-3B on seen targets.}
  \label{fig:3seen}
\end{figure}

\begin{figure}[htbp]
  \centering
\includegraphics[width=0.9\textwidth]{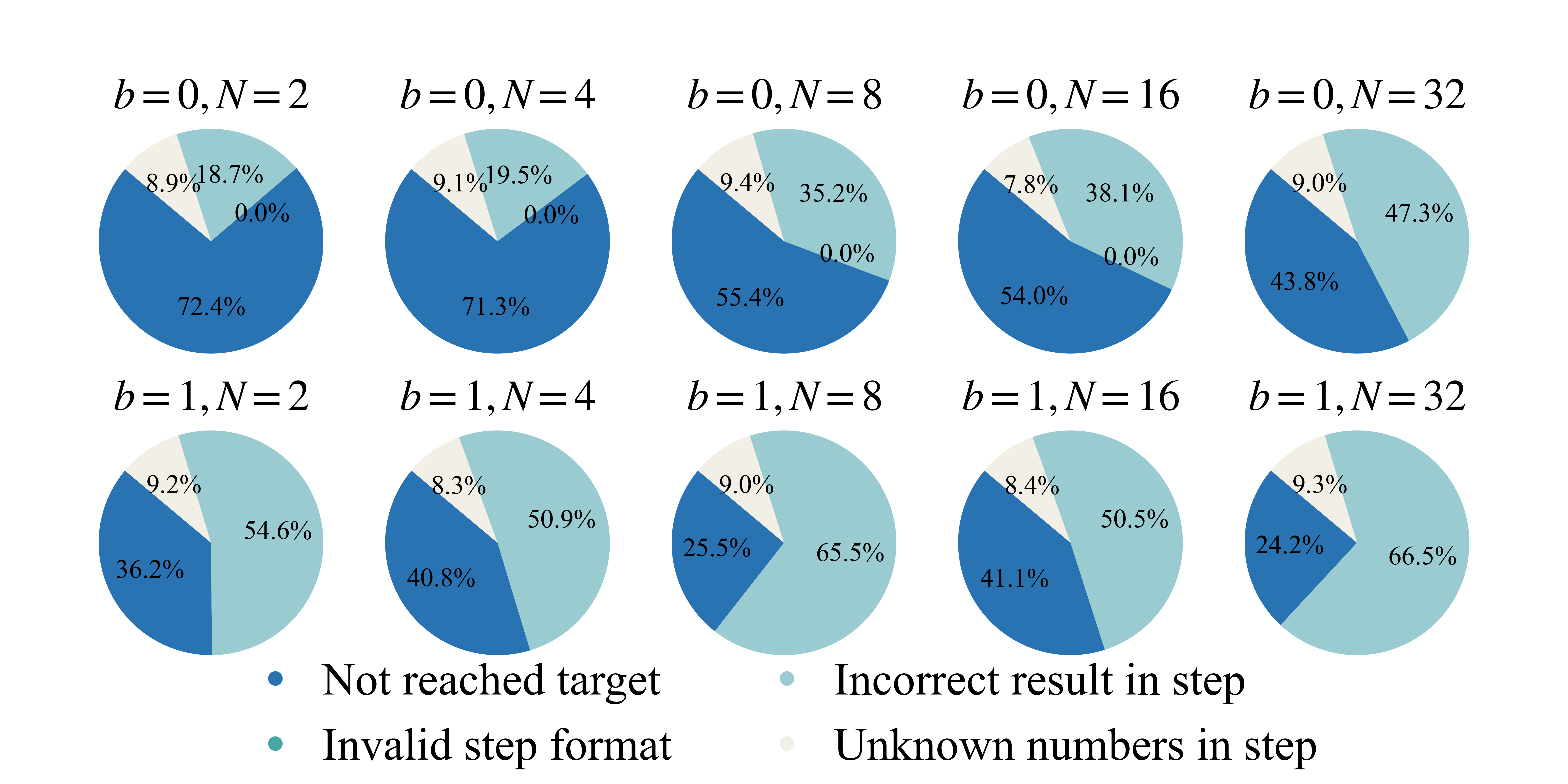}
  \caption{Error types of Llama3.2-3B on new targets.}
  \label{fig:3new}
\end{figure}

We demonstrate a more comprehensive analysis of error types, with results for Llama3.2-1B on Seen Targets shown in \ref{fig:1seen}, results on New Targets in \ref{fig:1new}, results for Llama3.2-3B on Seen Targets shown in \ref{fig:3seen}, and results on New Targets shown in \ref{fig:3new}.
Through empirical observation, we identify that the Llama3.2-3B model demonstrates superior computational accuracy. However, it frequently commits errors that prevent it from reaching the target. We hypothesize that this phenomenon is attributable to the model's robust foundational capabilities, which prioritize computational precision. A potential solution to mitigate this issue could involve the incorporation of more training data.

\subsection{Analysis for Temperatures}
\label{app:temp}
\begin{figure}[htbp]
  \centering
\includegraphics[width=0.8\textwidth]{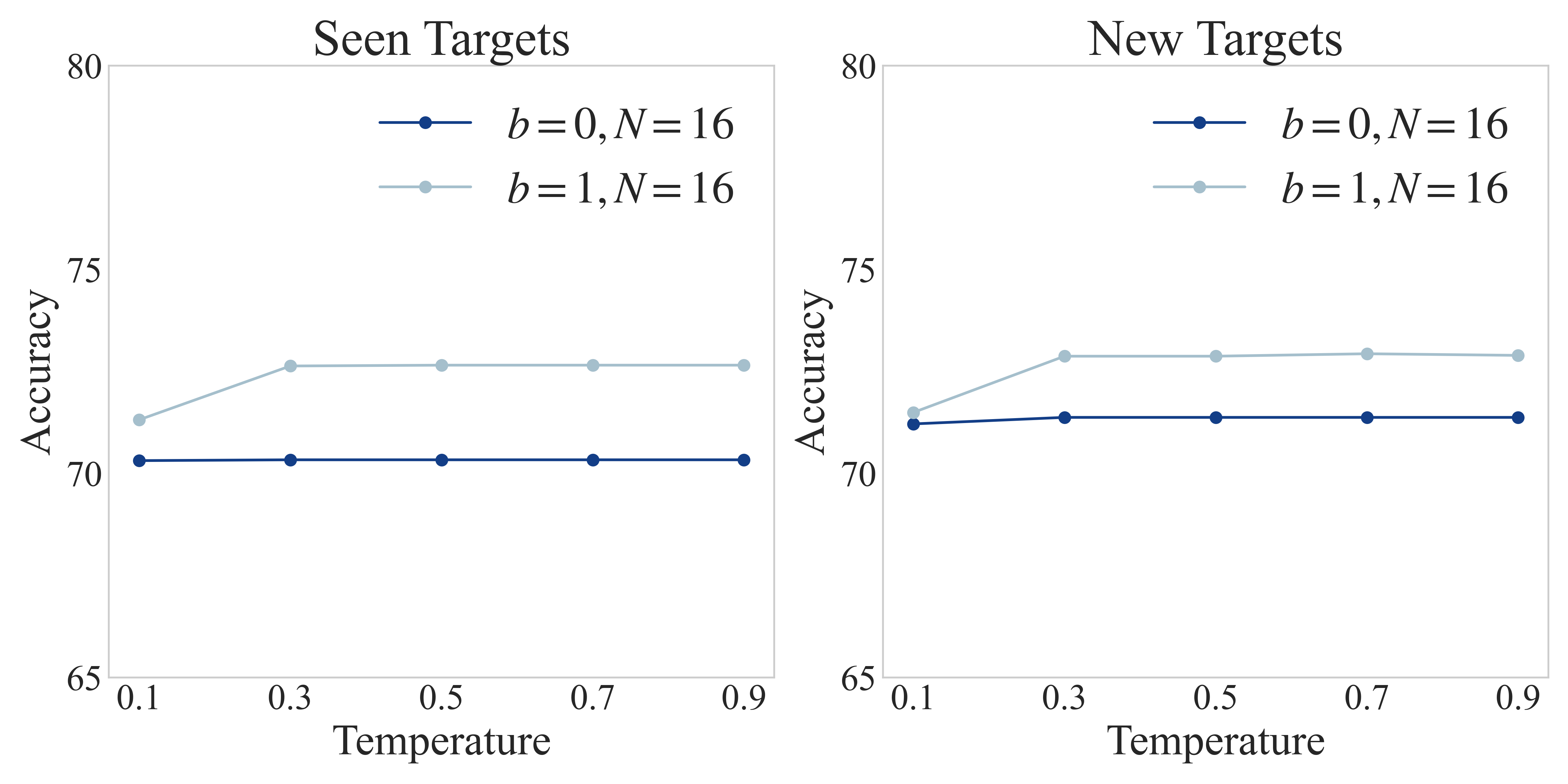}
  \caption{The influence of different temperatures on accuracy.}
  \label{fig:temperature}
\end{figure}

 In previous experiments, we consistently set the temperature parameter to 0.7. To further investigate the sensitivity of our algorithm to temperature variations, we conduct additional experiments, with the results presented in Figure \ref{fig:temperature}. The findings indicate that our method exhibits strong stability across different temperature settings, with only excessively low temperatures causing minor adverse effects specifically under the condition of $b=1$. Consequently, we recommend using the conventional temperature value of 0.7, which proves to be a reasonable choice.

\subsection{Analysis for Different Ratio Between $\mathcal{D}_{op}$ and $\mathcal{D}_{back}$}
We supplement the experimental results on the test split of New Targets across varying ratios between $\mathcal{D}_{op}$ and $\mathcal{D}_{back}$ using the Llama3.1-1B model in Table \ref{neg3b}.

\begin{table}[htbp]
\centering
\caption{Performance evaluation of self-backtraking under varying ratios of $\mathcal{D}_{op}$ to $\mathcal{D}_{back}$ for Llama3.2-1B on New Targets.}
\label{neg3b}
\resizebox{0.5\columnwidth}{!}{ 
\begin{tabular}{lccccc}
\toprule
\textbf{Parameter} & \textbf{1 : 0.5} & \textbf{1 : 1} & \textbf{1 : 2} & \textbf{1 : 3} & \textbf{1 : 4} \\
\midrule
$b=0, N=8$        & \textbf{68.58}  & 67.40  & 65.16  & 63.36  & 63.22  \\
$b=1, N=8$        & \textbf{69.36}  & 68.02  & 65.06  & 63.84  & 64.38  \\
$b=0, N=16$       & \textbf{71.98}  & 71.42  & 68.22  & 67.64  & 67.70  \\
$b=1, N=16$       & 72.64  & \textbf{72.94}  & 65.80  & 66.76  & 69.76  \\
\bottomrule
\end{tabular}}
\end{table}
\end{document}